\documentclass[10pt,twocolumn,letterpaper]{article}

\usepackage{cvpr}

\usepackage{float}
\usepackage{array}
\usepackage{colortbl}

\definecolor{firstcolor}{rgb}{1, 0.6, 0.6}
\definecolor{secondcolor}{rgb}{1, 0.8, 0.6}
\definecolor{thirdcolor}{rgb}{1,1, 0.6}
\newcommand{\fst}[1]{\cellcolor{firstcolor}#1}
\newcommand{\snd}[1]{\cellcolor{secondcolor}#1}
\newcommand{\trd}[1]{\cellcolor{thirdcolor}#1}

\definecolor{cvprblue}{rgb}{0.21,0.49,0.74}
\usepackage[pagebackref,breaklinks,colorlinks,allcolors=cvprblue]{hyperref}

\title{\methodname: 3D-to-2D Scene Diffusion Cascades for Urban Generation}
\author{Hanlei Guo$^{1}$, Jiahao Shao$^{1}$, Xinya Chen$^{1}$, Xiyang Tan$^{3}$, Sheng Miao$^{1}$, Yujun Shen$^{2}$, Yiyi Liao$^{1}$\thanks{Corresponding author.}\\ 
\vspace{-0.2cm}
\normalsize $^{1}$ Zhejiang University\quad$^{2}$ Ant Group\quad$^{3}$ The University of British Columbia\\
\small{Project Page:\url{https://xdimlab.github.io/ScenDi_website/}}
}

\newcommand{\bI}{\mathbf{I}}

\newcommand{\bs}{\mathbf{s}}
\newcommand{\bM}{\mathbf{M}}

\newcommand{\bR}{\mathbf{R}}

\newcommand{\bD}{\mathbf{D}}
\newcommand{\bz}{\mathbf{z}}

\newcommand{\bff}{\mathbf{f}}
\newcommand{\bF}{\mathbf{F}}
\newcommand{\bo}{\mathbf{o}}
\newcommand{\bO}{\mathbf{O}}
\newcommand{\bc}{\mathbf{c}}

\newcommand{\bepsilon}{\boldsymbol{\epsilon}}

\newcommand{\nR}{\mathbb{R}}

\newcommand{\cG}{\mathcal{G}}

\newcommand{\cL}{\mathcal{L}}

\newcommand{\cQ}{\mathcal{Q}}
\newcommand{\cC}{\mathcal{C}}
\newcommand{\cD}{\mathcal{D}}

\newcommand{\cE}{\mathcal{E}}

\newcommand{\cV}{\mathcal{V}}

\makeatletter
\DeclareRobustCommand\onedot{\futurelet\@let@token\@onedot}
\def\@onedot{\ifx\@let@token.\else.\null\fi\xspace}

\makeatother

\renewcommand{\eqref}[1]{Eq.~\ref{#1}}

\newcommand{\boldparagraph}[1]{\vspace{0.15cm}\noindent{\bf #1:} }

\newcommand{\gray}[1]{\noindent{\color{gray}{#1}}}

\newcommand{\methodname}{ScenDi}
\newcommand{\METHOD}{\texttt{\methodname}\xspace}

\newcolumntype{P}[1]{>{\centering\arraybackslash}m{#1}}

\begin{document}

\twocolumn[{%
\renewcommand\twocolumn[1][]{#1}%
\maketitle
\begin{center}
  {
  \vspace{-30pt}
  \captionsetup{type=figure}
  \centering
    \includegraphics[width=\linewidth]{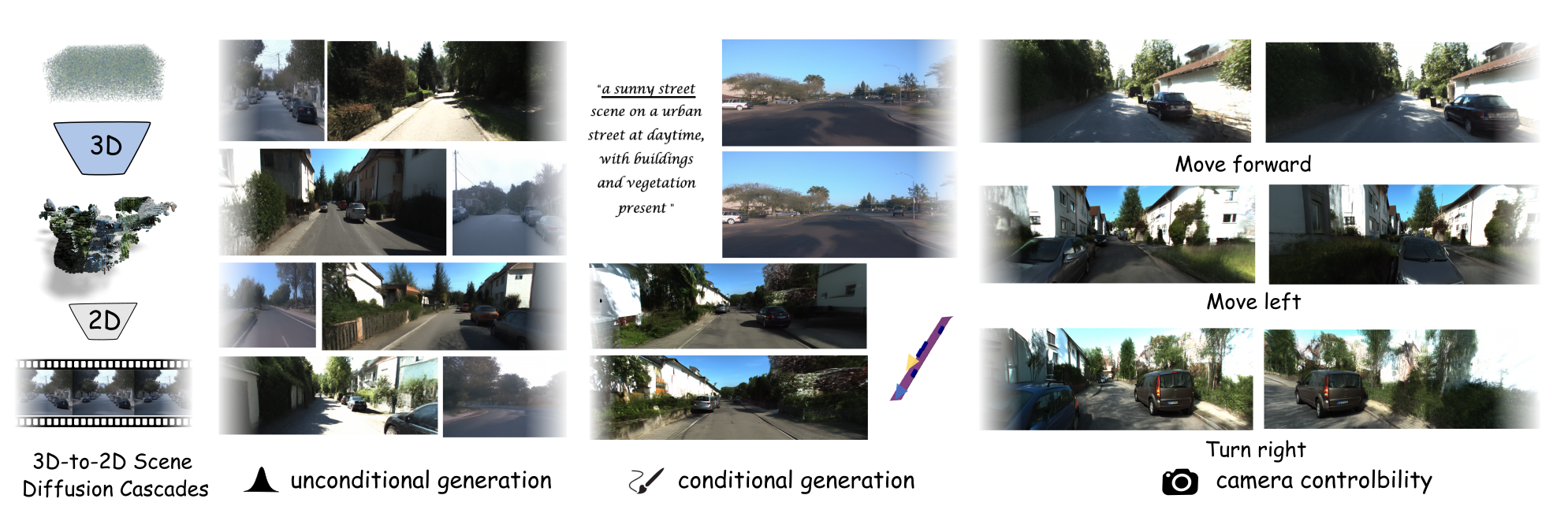}  
  \vspace{-15pt}
  \captionof{figure}{
  \textbf{\METHOD{}} generates high-quality urban scenes using a 3D-to-2D Scene Diffusion cascade, with optional condition signals like text and layout for controllable 3D space generation. Our method provides flexible camera control, even though our training data primarily consists of forward-moving trajectories.
  }
  \label{fig:teaser}
}
\end{center}
}]
\begin{abstract}
Recent advancements in 3D object generation using diffusion models have achieved remarkable success, but generating realistic 3D urban scenes remains challenging. Existing methods relying solely on 3D diffusion models tend to suffer a degradation in appearance details, while those utilizing only 2D diffusion models typically compromise camera controllability. To overcome this limitation, we propose ScenDi, a method for urban scene generation that integrates both 3D and 2D diffusion models. We first train a 3D latent diffusion model to generate 3D Gaussians, enabling the rendering of images at a relatively low resolution. To enable controllable synthesis, this 3DGS generation process can be optionally conditioned by specifying inputs such as 3d bounding boxes, road maps, or text prompts. Then, we train a 2D video diffusion model to enhance appearance details conditioned on rendered images from the 3D Gaussians. By leveraging the coarse 3D scene as guidance for 2D video diffusion, ScenDi generates desired scenes based on input conditions and successfully adheres to accurate camera trajectories. Experiments on two challenging real-world datasets, Waymo and KITTI-360, demonstrate the effectiveness of our approach.
\end{abstract}

\section{Introduction}
Generating 3D urban scenes from scratch, either unconditionally or with coarse guidance such as text prompts or layout maps, is an important step toward building open-world virtual environments for gaming and driving simulation.
Unlike image-to-video synthesis~\cite{vista,gen3c}, which extends an input view given viewpoints, 3D scene generation aims to create entire environments with consistent geometry and realistic appearance that can be rendered from free viewpoints.
Such capability is crucial for enhancing data diversity, supporting scene composition, and enabling scalable scene generation.

A majority line of work addresses this task by directly performing generation in the 3D space~\cite{urbangiraffe,nfldm,cc3d,infinicity}.
While such approaches offer explicit spatial modeling, their performance is constrained by the low resolution of 3D representations.
Besides, generating 3D urban scenes that directly render to high-fidelity images is challenging due to the scarcity of such 3D GT data.
More recently, another line of methods \cite{streetscapes, infinicube, uniscene, lu2024wovogen, consistentcity} attempt to enhance the performance by first generating 3D semantic voxels, and then render 2D conditions (e.g., depth or semantic maps) for video generation, allowing for producing high-fidelity details thanks to the powerful pre-trained 2D models.
While this approach benefits from 3D structure, the final appearance is still entirely generated in 2D, requiring the model to learn a complex mapping from depth or semantics to RGB images, which can make training less efficient and lack consistency when the same place is revisited.
This raises a fundamental question: How much of the generation process should occur in 3D space, and how much should be delegated to 2D?

In this work, we propose \METHOD{}\footnote{\METHOD{} means ``descend" in Italian, reflecting our cascaded generation process moving from 3D to 2D.}, a cascaded 3D-to-2D \textit{Scen}e \textit{Di}ffusion framework that leverages the complementary strengths of both modalities. Our key insight is that 3D generation should take the lead in establishing geometric and coarse appearance priors, while 2D diffusion focuses on refining details and enhancing distant regions. 
With this cascaded design, \METHOD{} achieves high-fidelity urban scene generation without 
sacrificing camera controllability, outperforming the counterpart that generates appearance purely in 2D space conditioned on geometry in training efficiency and loop consistency.

Our method begins by training a 3D latent diffusion model to generate coarse 3D scenes, which comprises a novel Voxel-to-3DGS VQ-VAE learned from 2D supervision and a 3D diffusion model, leveraging an off-the-shelf depth estimator to form the input voxel grids.
This enables sampling 3D Gaussian Splatting (3DGS) scenes that renders to multi-view consistent images.
However, rendered images from the 3D LDM often lack high-frequency details due to limited 3D resolution and fail to capture distant regions. To address this, we condition a 2D video diffusion model on the rendered RGB images, leveraging its ability to refine details and synthesize regions beyond predefined ranges. Experimental results on real-world autonomous driving datasets KITTI-360 and Waymo demonstrate that our 3D-to-2D diffusion cascades allow for generating high-fidelity urban scenes while preserving accurate camera control.

\section{Related Work}
 
\boldparagraph{Direct 3D Content Generaion}
One line of research focus on generating 3D scenes relying solely on 3D backbones. Early works \cite{eg3d, cc3d, citydreamer, urbangiraffe, discoscene} that adopt GANs \cite{Gan} suffer from inherent issues such as training instability and model corruption. A growing trend to leverage 3D diffusion models in image on tasks emerged due to their superb training stability. Methods lying on this line enables the availability of explicit 3D representation, thus leads to the camera controllable viewpoints over the generated scenes. Some works \cite{l3dg,sat2scene,blockfusion,diffgs,clay} construct 3D ground-truth datasets and perform training directly in 3D space. However, the natural scarcity of high-quality 3D scene data, such as 3D Gaussians that require per-scene optimization, undermines these works' scalability and limits their ability to generalize to diverse real-world scenarios. In contrast, we propose to use a VQ-VAE that directly maps voxel grids obtained from off-the-shelf depth priors to 3D Gaussians. Similar to us, several methods \cite{renderdiffusion,dmv3d,ssdnerf,nfldm,gaudi} also train 3D diffusion models that predict 3D representations with 2D multi-view image supervisions. Although they can avoid heavy dataset pre-processing, their results often lose fine appearance details. Our method addresses this challenge by using 3D-to-2D diffusion cascades.
The last line of methods~\cite{lin2023magic3d,chen2023fantasia3d,tang2023dreamgaussian,vsd,scenecraft} distills information from pretrained 2D diffusion models to optimize 3D representation based on Score Distillation Sampling (SDS)~\cite{dreamfusion}. These methods require a long training time to generate a single scene and usually suffer from over-saturation.

\boldparagraph{Multi-view Image Generation}
In contrast to direct 3D generation, some works leverage 2D diffusion models for multi-view generation, aiming to exploit the powerful generative prior of 2D diffusion models trained on internet-scale data. 
Despite the impressive progress in this direction, many methods target only on object-level generation~\cite{zero123plus,zero1to3, instant3d,syncdreamer, wonder3d,sv3d}.
There are also attempts for street scene image/video generation~\cite{magicdrive,drivedreamer,panacea,vista}. While most methods show superior visual fidelity compared with 3D generative methods, they struggle to provide flexible camera control due to the lack of explicit 3D information.
Recently, there has been a growing trend towards gradually reconstructing 3D scenes while generating novel views ~\cite{luciddreamer, wonderjourney, wonderworld}. However, these methods have not yet demonstrated the ability to handle large forward camera motions typical in driving scenarios.

\boldparagraph{Urban Scene Generation}
A line of methods~\cite{vista, panacea, drivewm, gen3c, autoscape, gem, drivearena} tackles image-to-video generation for urban scenes. Despite promising results, noticeable visual degradation often occurs as the methods extend far from the initial viewpoint. Another set of approaches~\cite{drivedreamer, gao2024magicdrive3d} uses coarse-condition guidance, such as HD maps or bounding boxes, to improve control over the generated scenes. However, these lack an explicit 3D backbone, leading to inter-frame inconsistencies. In contrast, our approach integrates diverse control signals for flexible scene generation while maintaining an explicit 3DGS backbone.
The most related methods~\cite{lu2024wovogen, streetscapes, infinicube, uniscene, consistentcity, lsd3d} generate a semantic voxel grid and rely on 2D renderings like depth or semantic maps for appearance synthesis. In comparison, our method directly generates coarse 3D Gaussians and refines them with a 2D model, achieving better training efficiency and improved loop consistency.

\section{Method}
\label{sec:method}
\begin{figure*}[t]
  \centering
  \includegraphics[width=\linewidth]{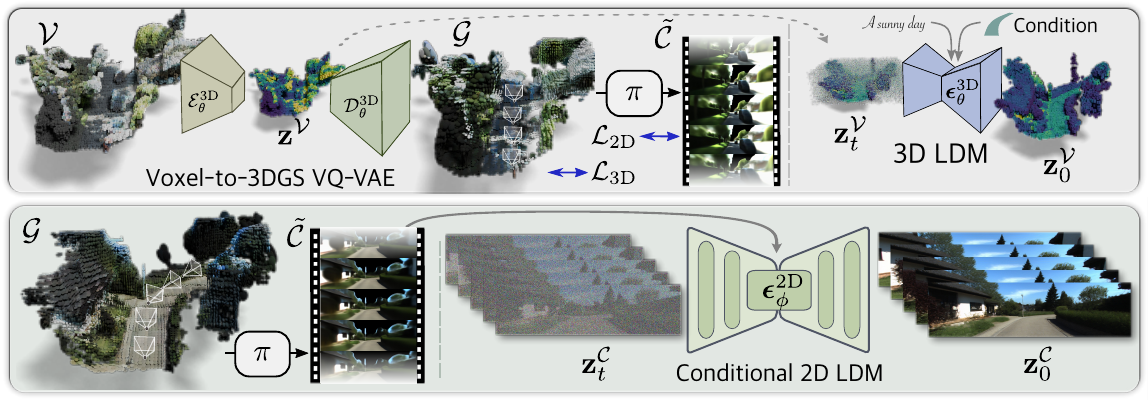}
   \caption{\textbf{Method Overview.} \METHOD{} leverages 3D and 2D diffusion cascades to generate high-quality urban scenes. \textit{Top:} We first build a Voxel-to-3DGS VQ-VAE to reconstruct scenes in a feed-forward manner. The input is a colored voxel grid $\cV$ constructed based on off-the-shelf metric depth estimator, whereas the output is a set of 3D Gaussian primitives $\cG$. Then, we train a 3D diffusion model  $\bepsilon_\theta^{\text{3D}}$ on the latent space $\bz^{\cV}$ to generate coarse 3D scenes, optionally conditioning on signals such as road maps or text prompts to enable explicit control over the content. \textit{Bottom:} Based on the coarse 3D scene, we train a 2D video diffusion model to refine foreground appearance details as well as generate distant areas. We achieve this by adopting video clip $\tilde{\cC}$ rendered from generated 3DGS as 3D conditional signals to fine-tune a conditional 2D latent diffusion model $\bepsilon_\phi^{\text{2D}}$.}
   \label{fig:pipeline}
   \vspace{-0.4cm}
\end{figure*}

\METHOD{} aims to generate high-quality and controllable urban scenes by incorporating both 3D and 2D diffusion models. An overview of our method is presented in \cref{fig:pipeline}, consisting of 3D generation and 2D refinement stages. Our 3D generator is a 3D Latent Diffusion Model, aiming to generate coarse geometry and appearance within a pre-defined volume as detailed in \cref{sec:3d_diffusion}. Here, we adopt 3D Gaussian primitives~\cite{3dgs} as the output 3D representation for its high rendering efficiency. Additionally, this generation process can be optionally controlled by 3D bounding boxes, road maps, and text instructions.  Next, we adopt a 2D video diffusion to fill in the details of rendered images from the 3D generation stage and synthesize background contents outside the volume as in \cref{sec:2d_diffusion}. The training process is elaborated in \cref{sec:training}.

\subsection{Scene Generation by 3D Diffusion}
\label{sec:3d_diffusion}

Following the standard LDM~\cite{ldm}, we design a 3D LDM for generating 3D Gaussian primitives, consisting of a 3D VQ-VAE and a latent-space diffusion model. While there are existing works in this direction for generating occupancy or semantic voxel grids~\cite{semcity, pdd, xcube, blockfusion,lt3sd}, these methods require accurate 3D GT data, i.e. the GT semantic voxel grids are used as input and output of the VQ-VAE. However, it is extremely time-consuming to obtain a large amount of 3D Gaussians for urban scenes using per-scene optimization. Therefore, we propose to learn a voxel-to-3DGS VQ-VAE directly from 2D supervision and pseudo 3D GT obtained from metric monocular depth estimations.

\boldparagraph{Input Voxel Grid Construction}
We first construct a colored voxel grid $\cV\in \nR^{H \times W \times D \times 3}$ as input to our 3D VQ-VAE. This is achieved by leveraging an metric monocular depth estimator. Specifically, given several posed input images $\left\{\bI_{i}\in\nR^{h \times w \times 3}\right\}^{N}_{i=1}$ within a volume of interest, we use a pre-trained depth estimator~\cite{metric3d} 
to infer their depth maps $\left\{\bD_{i}\in \nR^{h \times w}\right \}^{N}_{i=1}$. Next, we unproject the depth maps and merge them into a unified coordinate system based on their camera poses. 
We follow~\cite{ucnerf} to create a consistency check module to examine inaccurate depth predictions and adopt filters to remove outliers. 
By merging information from different camera views, we obtain a global RGB point cloud 
which provides a coarse 3D scene context. This colored point cloud is then discretized into the RGB voxel grid $\cV$, where points outside of the voxel grid are discarded.
We refer to $\cV$ as the foreground, which only covers a limited range of the unbounded urban scene.We project the merged point cloud into each input view to obtain corresponding foreground masks $\left\{\bM_{i}\right\}^{N}_{i=1}$, such that our VQ-VAE only needs to reconstruct scenes within the foreground region. \par

\boldparagraph{Voxel-to-3DGS VQ-VAE} Taking the voxel grid $\cV$ as input, we train a VQ-VAE which maps it into a lower dimensional latent code $\bz\in\nR^{H' \times W' \times D' \times F'}$ and then decodes it into a set of 3D Gaussian primitives $\cG$. More formally, let $\cE_\theta^{3D}$ denotes the encoder,  $\cQ_\theta^{3D}$ the vector quantizer, and $\cD_\theta^{3D}$ the decoder of the 3D VQ-VAE, we map the input $\cV$ to the Gaussian primitives $\cG$ as follows:
\begin{equation}
\bz^{\cV} = \cE_\theta^\text{3D}(\cV), \;{\bz^{\cV}_q} = \cQ_\theta^\text{3D}(\bz^{\cV}), \;\cG = \cD_\theta^\text{3D}(\bz^{\cV}_q)
\label{eq:vq_vae}
\end{equation}
In practice, our decoder $\cD_\theta^{3D}$ predicts scene occupancy $\bO\in R^{H \times W \times D}$ through an occupancy branch and a feature volume $\mathbf{F}\in R^{H \times W \times D \times F}$ at the same spatial resolution of $\cV$ through another feature branch.
We denote one Gaussian primitive through MLP heads when a voxel is predicted as occupied. %
Each 3D Gaussian primitive is parameterized by the following attributes: color $\bc$, opacity $\alpha$, scale $\bs$, rotation $\bR$, and an offset to the voxel center $\Delta \bo$.  %
Let $f_\theta^\text{color}$, $f_\theta^\text{opa}$, $f_\theta^\text{geo}$, and $f_\theta^\text{offset}$ denote the MLP heads for predicting the Gaussian attributes, and $\bff\in \nR^{F}$ denotes a feature vector responding to one voxel of $\bF$, we obtain the attributes of one 3D Gaussian primitive from an occupied 3D voxel as follows:
\begin{align}
\bc = f_\theta^\text{color}(\bff), &~~~ \alpha =f_\theta^{\text{opa}}(\bff), \\
\bs, \bR = f_\theta^{\text{geo}}(\bff) , &~~~ \Delta \bo =f_\theta^{\text{offset}}(\bff)
\end{align}
The 3D Gaussian primitives can then be rendered into images through the rendering function $\pi$.

\boldparagraph{Latent 3D Diffusion Model}
After training a generalizable feed-forward reconstruction network, we train a 3D diffusion model that operates on the 3D latent space of the VQ-VAE, which allows for generating a 3D latent code from pure noise.

More specifically, we follow the standard forward process of the diffusion model by adding noise to the original clean latent $\bz_0$.
\begin{equation}
\bz^{\cV}_t = \sqrt{\overline{\alpha}_t}\bz^{\cV}_0 + \sqrt{1-\overline{\alpha}_t}{\bepsilon},\;{\bepsilon}\in  \mathcal{N}(0,1)
\label{eq:forward}
\end{equation}
where $\overline{\alpha}_{t}$ controls the level of noise added to the original latent and $\bz_t$ is the noisy version of $\bz_0$ at time step t.
For the reverse process, we train a noise estimator $\bepsilon_\theta(\bz_t; t)$ for denoising, which follows the standard DiT~\cite{dit} network architecture for image generation~\cite{ddpm}.
In practice, we train our diffusion model to perform v-prediction~\cite{v_predict}. The predicted clean latent $\hat{\bz}_0$ can be derived from the direct output of the network by:
\begin{equation}
    \hat{\bz}^{\cV}_0 = \sqrt{\overline{\alpha}_t}\bz^{\cV}_{t} - \sqrt{1-\overline{\alpha}_t}\bepsilon_\theta^{3D}(\bz^{\cV}_{t};t, c)
\label{eq:v_predict}
\end{equation}
where $c$ indicates optional condition signals described below. During sampling, the model denoises a pure Gaussian noise $\epsilon$ over multiple iterations to a clean latent using a standard DDIM sampler~\cite{ddim}. 

\boldparagraph{Controllable 3D Scene Generation} To improve controllability, we incorporate optional conditional signals during training of the 3D latent diffusion model. We explore two settings: coarse 3D layout conditioning and text conditioning.
For layout conditioning, our 3D generative model directly supports 3D constraints without rendering them into 2D space~\cite{drivedreamer, lsd3d}. Specifically, we build a 3D conditional volume from primitive labels~\cite{kitti360}, where each voxel contains a one-hot semantic label (e.g., vehicle, road, or empty). This volume is downsampled to the latent resolution and concatenated with the initial latent $\bz_0$ for diffusion training.
For text-conditioned generation, we follow ~\cite{cosmosreason} to annotate each scene with attributes such as weather, road type, and background, forming a descriptive sentence. Text embeddings extracted via ~\cite{clip} are then injected into the model through cross-attention.

\subsection{Scene Augmentation by 2D Diffusion}
\label{sec:2d_diffusion}
Although images synthesized by 3D LDM excel in maintaining cross-view consistency and disentangling camera poses from contents, they may suffer from blurriness due to the resolution limitation. Additionally, far areas are difficult to model in 3D. Therefore, we use the prior information of the pre-trained 2D model~\cite{svd, wan2025} to augment appearance details while completing distant areas.

\boldparagraph{Conditional Latent 2D Diffusion Model}
Inspired by recent diffusion-based super-resolution methods~\cite{supir, stablesr}, we use a 2D video diffusion model to perform conditional generation.
Let $\cC = [\bI_1,\cdots, \bI_K] \in \nR^{K\times h \times w \times 3}$ denotes a target RGB video clip consisting of $K$ frames, and $\tilde{\cC} = [\tilde{\bI}_1,\cdots,\tilde{\bI}_K] \in \nR^{K\times h \times w \times 3}$  denotes the corresponding video produced by our Voxel-to-3DGS VQ-VAE. We train a 2D video diffusion model for generating $\cC$ conditioned on $\tilde{\cC}$.
Here, we also adapt a latent video diffusion model, using a frozen VAE from~\cite{svd, wan2025} to map the original videos to the latent space. Let
$\bz^{\tilde{\cC}}$ denotes the latent of rendered video $\tilde{\cC}$, and $\bz^{\cC}$ denotes to latent of final refined video ${\cC}$. We perform the standard forward process by adding noise to $\bz^{\cC}$, same as in \cref{eq:forward}. For the reverse process, we learn a conditional noise estimator $\bepsilon_\phi(\bz^{\cC}_t; t, \bz^{\tilde{\cC}})$.

\boldparagraph{Training} During training, we use images synthesized from VQ-VAE reconstruction and their ground-truth images as paired data and project both type of images into latent space via frozen VAE encoder $\cE_\phi^{2D}$. Inspired by~\cite{marigold, chronodepth}, We adopt channel-concatenation as our condition mechanism. Specifically, we only apply random noises to $\bz^{\bI_{gt}}$ following equation~\eqref{eq:forward} to obtain its noisy version $\bz_t^{\cC}$ and concatenate it with $\bz^{\tilde{\cC}}$ to formulate the final input to U-Net. After fine-tuning the original model, we can denoise noisy latent into a high-quality video clip under the guidance of coarse 3D prior $\tilde{\cC}$.\par
\boldparagraph{Inference} To address the computational demands of training a 2D video diffusion model, we limit training to a small subset of frames. %
This yields abrupt transitions between clips when applied to longer video sequences with a simple replacement trick~\cite{repaint}.
Inspired by~\cite{diffusionforcing,chronodepth}, we use Diffusion Forcing training and inference strategy. Specifically, after training with aforementioned setup, we finetune the network by independently sampling distinct noise levels for each individual frame within the clip instead of only one noise level for the whole clip. During sampling time, we condition previously generated $W$ frames to the later $F-W$ frames via concatenating along the time dimension. The time embedding for the whole clip is $\mathbf{t} = [\underbrace{t_{\epsilon}, t_{\epsilon}, ..., t_{\epsilon}}_{W}, \underbrace{t, t, ..., t}_{F-W}]$, where time embedding $t_{\epsilon}$ is a small timestep and $t$ is sampled from a fixed common diffusion scheduler. In this way, we can do sampling conditioned on previously generated frames.

\subsection{Training Losses} 
\label{sec:training}
The whole training process can be divided into three stages. First, we train a generalizable VQ-VAE to reconstruct multiple scenes. For geometry reconstruction, we use a simple BCE loss to regularize the predicted occupancy $\bO$
to be consistent with the occupancy of input voxel grid $\cV$.
For appearance reconstruction, we utilize L1 loss and SSIM loss following 3DGS~\cite{3dgs}, with a mask loss to separate foreground and background. %
We sample $M$ frames to render for each single scene to apply the 2D image loss, yielding the full loss for one scene as: 
\begin{align}
    \cL_{recon} & = \cL_{\text{3D}} + \sum_{m=1}^{M}\cL_{\text{2D}}^m \\
    \cL_{\text{3D}} & = \lambda_{bce}\cL_{bce} + \lambda_{vq}\cL_{VQ} \\
    \cL_{\text{2D}} & = \lambda_{rgb}\cL_1 + \lambda_{ssim}\cL_{ssim} + \lambda_{fg}\cL_{fg}
\end{align}
Secondly, we train the 3D diffusion model to generate coarse 3D scene latent. We compute Mean Squared Error between the clean sample $\bz^{\cV}_{0}$ and predicted clean sample $\hat{\bz}^{\cV}_{0}$ derived from $\cref{eq:v_predict}$ to supervise our model:
\begin{equation}
\cL_{diff}^{3D} = ||\bz^{\cV}_0 - \hat{\bz}^{\cV}_0||^2
\end{equation}
Finally, for 2D diffusion model fine-tuning, we following the training procedure of pre-trained models~\cite{svd, wan2025} as our training loss:
\begin{equation}
\cL_{diff}^{2D} = ||\bz^{\cC}_0 - \hat{\bz}^{\cC}_0||^2
\end{equation}
where $\hat{\bz}^{\cC}_0$ denotes the predicted original latent.

\section{Experiments}
\label{sec:exp}
\subsection{Implementation Details}
\label{sec:imple}
\boldparagraph{Datasets}
We conduct our experiments on two outdoor autonomous driving datasets KITTI-360 \cite{kitti360} and Waymo \cite{waymo}. For both datasets, we set $H=32$, $W=128$, $D=192$ for the voxel grid $\cV$,  corresponding to [$-3$m, $9.8$m] for height, [$-25.6$m, $25.6$m] for width, and [$-20$m, $56.8$m] for the car forwarding direction.
We filter out scenes with large turn, slow ego-motion, or significant dynamic objects. %
For our 2D loss $\cL_{\text{2D}}$ to train VQ-VAE, we render RGB images at the resolution of $128 \times 256$ and resize the GT to the same resolution. For the 2D diffusion fine-tuning stage, we scale up image resolution to $640\times 960$ for both datasets. Due to dataset scale limitation, we introduce spatial overlap in voxel grids, resulting in $\sim$20k samples for Waymo and $\sim$35k for KITTI-360. Further preprocessing details are provided in supplementary.

\boldparagraph{3D Diffusion} 
For our Voxel-to-3DGS VQ-VAE, we replace standard 3D convolutions with Asymmetrical Residual Blocks from Cylinder3D \cite{cylindrical} for efficiency. The encoder downsamples the input by a factor of 4. %
During training, we use $M=4$ 2D supervision images per scene for both datasets.
For the reconstruction loss, we set $\lambda_{bce}$ = 1,  $\lambda_{vq}$ = 0.25, $\lambda_{rgb}$ = 0.8, $\lambda_{ssim}$ = 0.2, $\lambda_{fg}$ = 0.5. Our 3D diffusion model is a 3D variant adapted from ~\cite{dit3d}, following the original transformer architecture of \cite{dit}. Note that we train on KITTI-360 and Waymo jointly and inject a dataset ID embedding for dataset awareness. 
VQ-VAE is trained from scratch for 300k steps using a batch size of 8 and latent 3D diffusion model is trained from scratch for 100k steps using a batch size of 32 separately with an initial learning rate of 1e-4. We use 8 A100 GPUS for training, our model requires approximately 2 days for VQ-VAE training and 3 days for 3D Diffusion training.

\boldparagraph{2D Diffusion}  
We trained two variants of the 2D diffusion model, each based on a different pretrained video diffusion backbone: Wan2.1-1.3B-i2v~\cite{wan2025} and SVD~\cite{svd}. This setup allows us to ensure a fair comparison with previous SVD-based methods~\cite{vista}, while also exploring the potential of our approach with the stronger WAN backbone. We initialize our models from the corresponding pretrained weights and reuse their VAEs, which remain frozen during training. For the SVD variant, following~\cite{chronodepth}, we sequentially fine-tune the spatial and temporal layers. First, we fine-tune spatial layers on single-frame images for 12k steps (batch size 32), then temporal layers on video clips for 25k steps (batch size 8), with randomly sampled clip length up to 5. For WAN-variant, we fine-tune transformer modules for 25k steps, with randomly sampled clip length up to 17. To enhance temporal consistency, we apply the diffusion forcing strategy \cite{diffusionforcing} on both models for an additional 15k steps. We use a learning rate of $3 \times 10^{-5}$ with AdamW on a cluster of 8 NVIDIA Tesla A100 GPUs for training. The SVD and Wan2.1-1.3B fine-tuning required approximately 5 days.

\begin{table*}[htbp]
\vspace{0pt}
  \centering
  \setlength{\tabcolsep}{4.2pt}
  \begin{tabular}{cccc|ccccc}
    \toprule
    \small
    Method  & Setting & Conditions & Backbone & FID$\downarrow$  & KID$\downarrow$ & FVD$\downarrow$  & TransErr$\downarrow$  & RotErr$\downarrow$ \\
    \midrule
    DiscoScene~\cite{discoscene}  & 3D Gen & 3D Bbox & 3D GAN& 135.3  & 0.093  & 2025.9  & 5.45 & 2.07 \\
    CC3D~\cite{cc3d} & 3D Gen &  BEV Layout & 3D GAN & 90.8 & 0.091  & 706.1  & 1.82  & 1.76 \\
    UrbanGen~\cite{urbangen} & 3D Gen & 3D Bbox+Semantic Voxel & 3D GAN & \snd{33.0}  & \snd{0.017}   &  \snd{300.1}  & \snd{0.21} & \trd{0.25} \\ 
    Ours & 3D Gen & 3D Bbox+Road Map & SVD~\cite{svd} & \trd{36.9}  & \trd{0.026}  & \trd{400.3}   & \trd{0.23} & \fst{0.13}  \\ 
    Ours & 3D Gen & 3D Bbox+Road Map & WAN2.1-1.3B~\cite{wan2025} & \fst{22.9}  & \fst{0.016}  & \fst{262.6}   & \fst{0.06} & \snd{0.23} \\ 
    \midrule
    ~\gray{Vista~\cite{vista}} & ~\gray{I2V} & ~\gray{Ref Images} &  ~\gray{SVD~\cite{svd}} & ~\gray{25.6} & ~\gray{0.016} & ~\gray{234.0} & ~\gray{2.33} & ~\gray{0.74} \\
    ~\gray{Gen3C~\cite{gen3c}}  & ~\gray{I2V} & ~\gray{Ref Images} & ~\gray{Cosmos~\cite{cosmos}} & ~\gray{24.1} & ~\gray{0.012} & ~\gray{426.1} & ~\gray{0.04} & ~\gray{0.11}  \\
    \bottomrule
  \end{tabular}
  \vspace{-0.6em} 
  \caption{\textbf{Quantitative Comparison} about video quality and camera controllability on KITTI-360 . 
  }
  \label{tab:quant_3d}

\vspace{-0.3cm}
\end{table*}
\begin{figure*}
    \centering
     \setlength{\tabcolsep}{0.5pt}
     \begin{tabular}{cccc}
    &\includegraphics[width=0.32\linewidth]{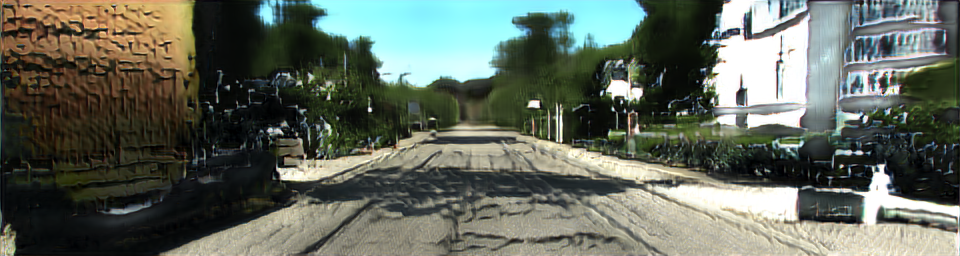} 
    &\includegraphics[width=0.32\linewidth]{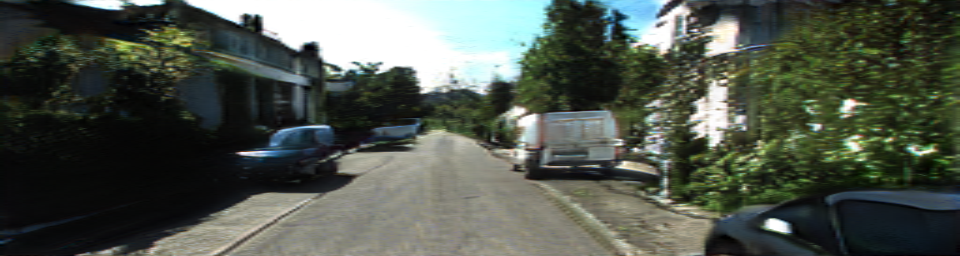} 
    &\includegraphics[width=0.32\linewidth]{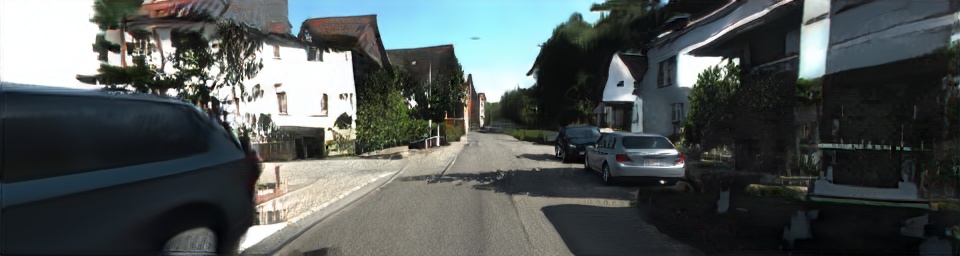}\\
    &\includegraphics[width=0.32\linewidth]{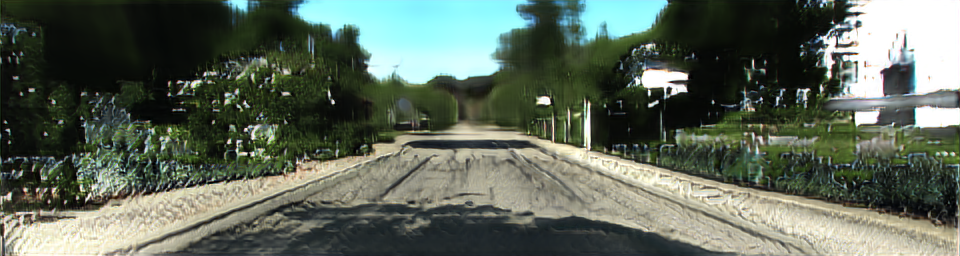} 
    &\includegraphics[width=0.32\linewidth]{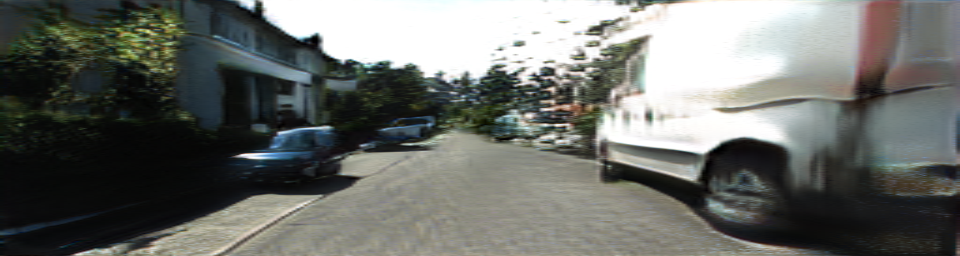} 
    &\includegraphics[width=0.32\linewidth]{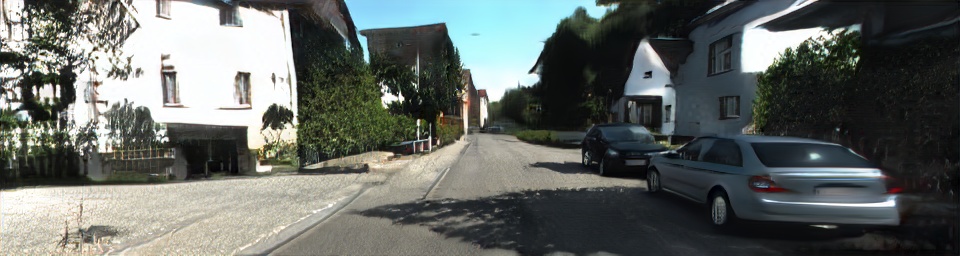} \\
    &\includegraphics[width=0.32\linewidth]{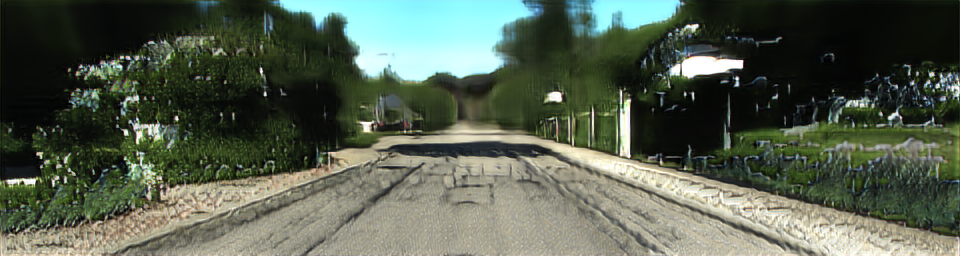}
    &\includegraphics[width=0.32\linewidth]{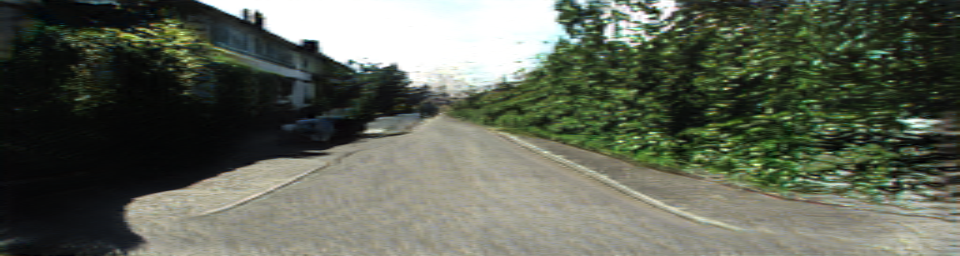} 
    &\includegraphics[width=0.32\linewidth]{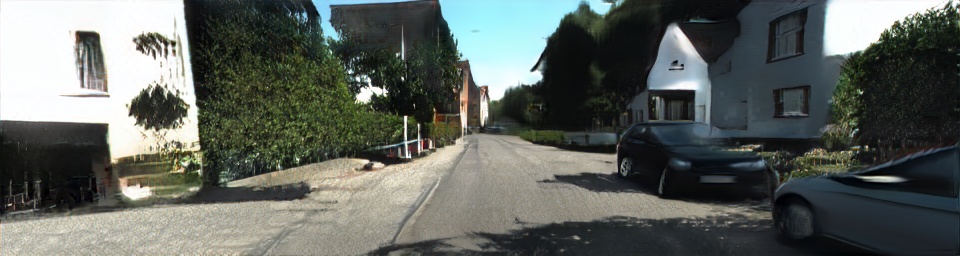} \\
     &CC3D~\cite{cc3d} &DiscoScene~\cite{discoscene} &UrbanGen~\cite{urbangen} \\
    &\includegraphics[width=0.32\linewidth]{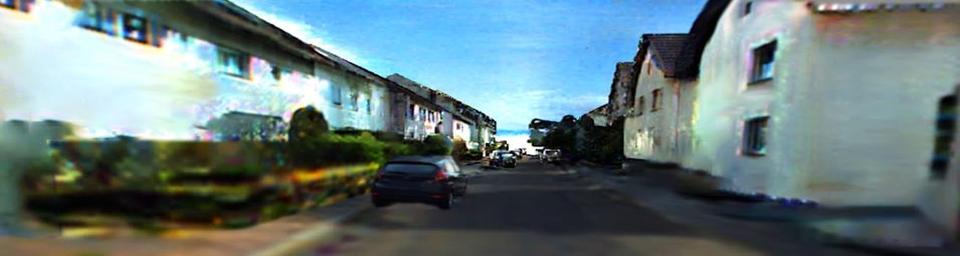} 
    &\includegraphics[width=0.32\linewidth]{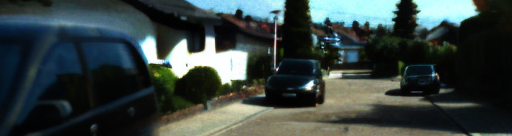} 
    &\includegraphics[width=0.32\linewidth]{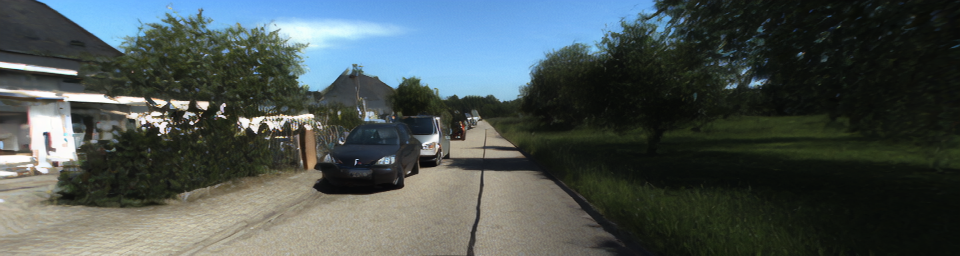} \\
    &\includegraphics[width=0.32\linewidth]{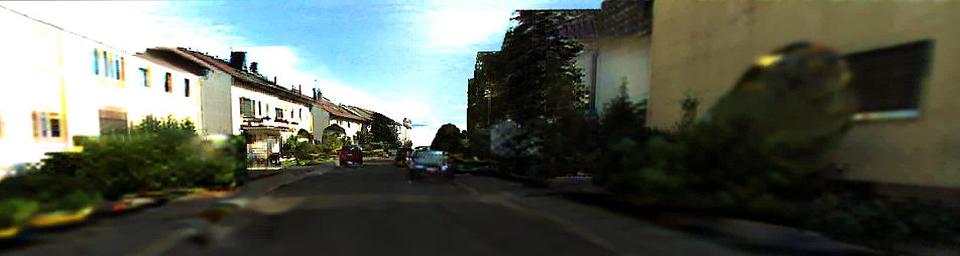} 
    &\includegraphics[width=0.32\linewidth]{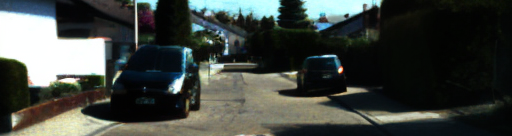} 
    &\includegraphics[width=0.32\linewidth]{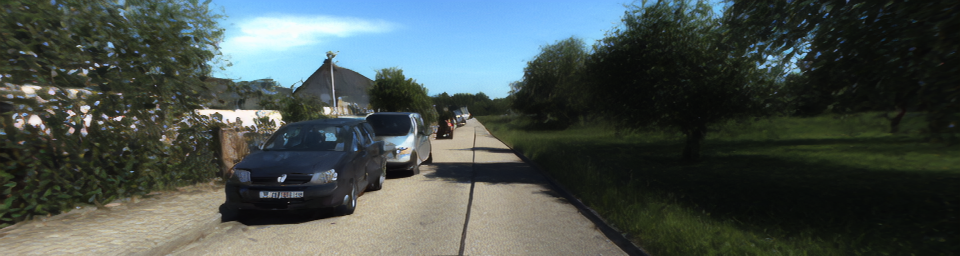}\\
    &\includegraphics[width=0.32\linewidth]{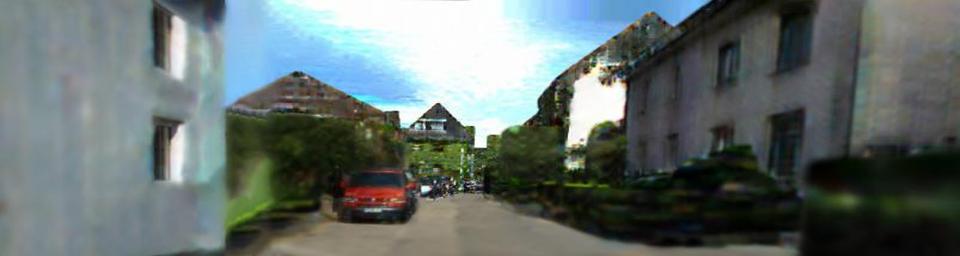} 
    &\includegraphics[width=0.32\linewidth]{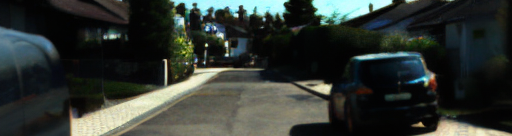} 
    &\includegraphics[width=0.32\linewidth]{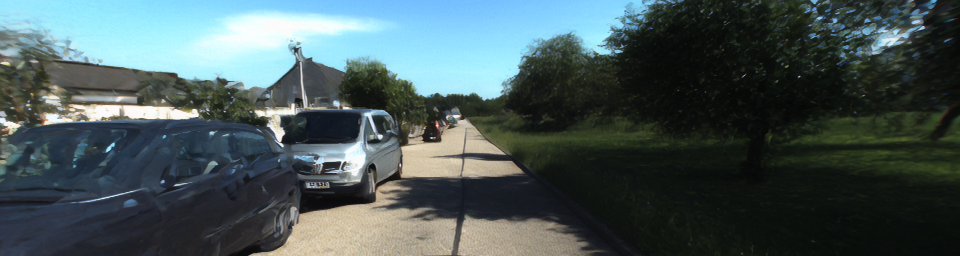}\\
    &GaussianCity~\cite{gaussiancity} &Urban Architect~\cite{urbanarchitect} &Ours \\
    \end{tabular}
    \caption{\textbf{Qualitative Comparison on KITTI-360.} We show results under forward moving situations. Our method achieve superior visual quality over other baselines.}
   \label{fig:kitti-quali}
   \vspace{-0.6cm}
\end{figure*}

\subsection{Baselines and Metrics} 
\boldparagraph{Metrics}
We evaluate the image quality of generated scenes by using commonly adopted metrics FID \cite{fid}, KID \cite{kid}. For FID and KID evaluation, we use 5k real samples and generated images to measure the distribution similarity. We further evaluate FVD~\cite{fvd} on 1k clips to quantify the visual quality and temporal consistency of the generated videos. We use
DUSt3R~\cite{dust3r} to estimate camera poses of the generated scenes and calculate the rotation error (RotErr) and translation error (TransErr) following ~\cite{cameractrl}. We highlight the \colorbox{firstcolor}{best}, \colorbox{secondcolor}{second-best}, and \colorbox{thirdcolor}{third-best} scores achieved on any metrics.

\boldparagraph{Baselines} We adopt the current state-of-the-art 3D urban generation methods as our baselines~\cite{cc3d, discoscene, urbangen,gaussiancity,urbanarchitect}. Recent image-to-video scene generation methods~\cite{vista,gen3c} are also included as a reference despite their inherent differences in target tasks.
DiscoScene~\cite{discoscene}, CC3D~\cite{cc3d}, UrbanGen~\cite{urbangen} and GaussianCity~\cite{gaussiancity} are GAN-based methods with different layout conditions: DiscoScene~\cite{discoscene} relies on foreground object layouts, CC3D~\cite{cc3d} uses BEV semantic maps, UrbanGen~\cite{urbangen} uses semantic voxel grids, and GaussianCity~\cite{gaussiancity} uses BEV semantic and height maps.
While DiscoScene, CC3D and UrbanGen use neural radiance fields as 3D representations, GaussianCity is a 3DGS-based method similar to us.
Different from these GAN-based methods, Urban architect~\cite{urbanarchitect} is based on a pre-trained diffusion model that optimizes an urban scene using VSD loss~\cite{vsd} given provided 3D layout priors. As for image-to-video baselines, we adopt Vista~\cite{vista} and Gen3C~\cite{gen3c}, where the former is fine-tuned from SVD~\cite{svd} conditioned on various control signals such as speed and steering angle, and the latter warps monocular depth to target views and inpaints it with a Cosmos~\cite{cosmos} model. 

\subsection{Quantitative and Qualitative Comparison}
\begin{figure}[t]
    \centering
    \centering
\setlength{\tabcolsep}{1pt}
\begin{tabular}{P{2cm}P{6cm}}
\includegraphics[width=0.3\columnwidth]{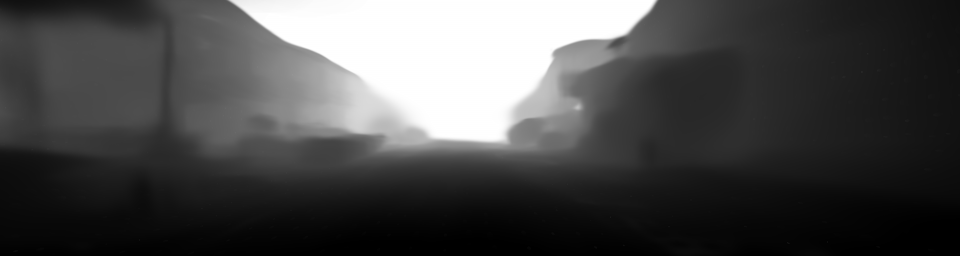} 
&\includegraphics[width=0.7\columnwidth]{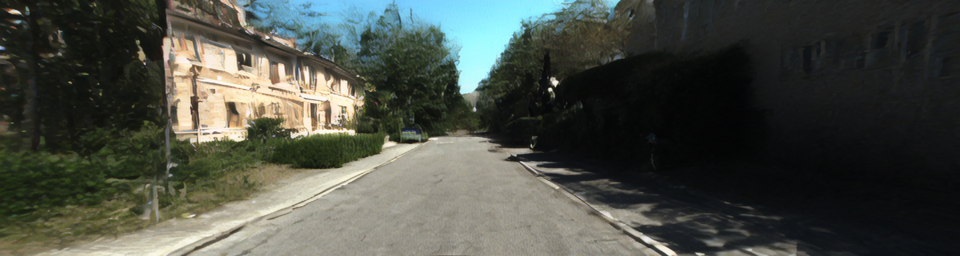} \\
\includegraphics[width=0.3\columnwidth]{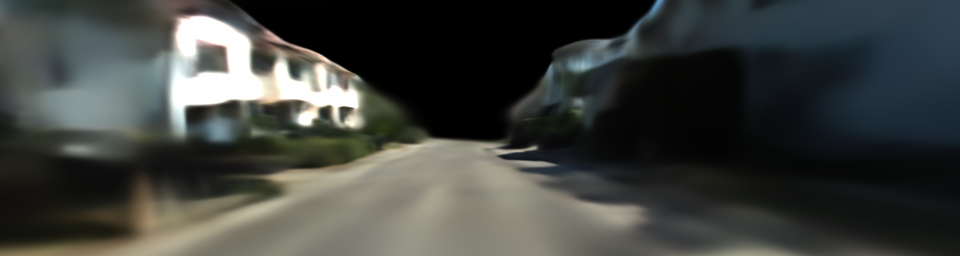} 
& \includegraphics[width=0.7\columnwidth]{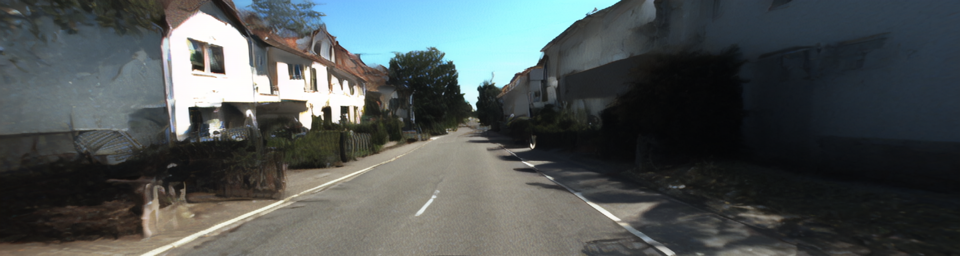} \\
\end{tabular}
 \vspace{-0.3cm}
\caption{\textbf{Ablation on Conditional Signal} for 2D augmentation. We show samples generated by different conditional signals on KITTI-360 after same training steps.}
\vspace{-0.3cm}
\label{fig:svd_ablation}

\end{figure}
\boldparagraph{Comparison with 3D Generation baselines}
We compare visual quality and camera controllability with 3D generative baselines~\cite{cc3d,discoscene,urbangen,gaussiancity,urbanarchitect} in~\cref{tab:quant_3d} and show qualitative comparisons in ~\cref{fig:kitti-quali}.
DiscoScene achieves reasonable results for foreground objects, but background generation may occasionally fail. CC3D~\cite{cc3d} and UrbanGen~\cite{urbangen} improve the background based on BEV or semantic voxel conditions, but their 3D feature volume representations have resolution constraints. GaussianCity~\cite{gaussiancity} provides a more compact scene representation by adopting BEV-Point representation but is trained from scratch on scene datasets, thus posing challenges in capturing high-frequency textures. Urban Architect~\cite{urbanarchitect} leverages diffusion model distillation with LG-VSD loss, which requires an extremely long processing time. Our method leverages the rich prior knowledge in the video diffusion model to synthesize more photorealistic appearances in the second augmentation stage, yielding the best FID, KID and FVD. We also designed various out-of-distribution trajectories involving translation and rotation to evaluate camera controllability of all methods. Since our method first employs 3D diffusion to generate explicit 3DGS, it achieves the same level of camera control accuracy as pure 3D generation methods. All 3D generation baselines are trained on forward-facing views, therefore large camera viewpoint changes can cause drastic image quality degradation.  DUSt3R may extract inaccurate poses in such cases, explaining the low camera accuracy of certain methods like DiscoScene and CC3D.
Since the layout data and training code are unavailable, we only present qualitative results on Gaussiancity and Urban Architect based on their paper or provided dataset samples.

\boldparagraph{Comparison with Image-to-video Generative baselines}
We also compare with image-to-video methods~\cite{vista, gen3c} and quantitative results are shown in ~\cref{tab:quant_3d}.
Since the generated frames from these methods often overlap with the GT frames, comparing FID/FVD is not entirely fair. However, our method exhibits a comparable performance. Notably, when the input and current viewpoints exhibit minimal overlap, Gen3C frequently suffers from significant quality degradation and inconsistency. This observation explains its relatively high FVD score, which is sensitive to performance drops within the video clip. To test camera controllability, we utilized the identical camera trajectory as the 3D generation baselines. We observe that pure 2D methods~\cite{vista} usually maintain a constant orientation on the straight road when conditioned on a small-angle curve signal while methods incorporating an explicit 3D representation achieve a more accurate camera viewpoint. We provide further details and qualitative results on this experiment in the supplementary.
\begin{table}[t]
    \resizebox{0.44\textwidth}{!}{
\centering
\begin{tabular}{l|ccc}
\toprule[1.2pt]
 & PSNR $\uparrow$ & SSIM $\uparrow$ & LPIPS $\downarrow$ \\
\midrule[1.2pt]
w/o $\cL_{bce}$, $G=1$ & \trd{21.47} & \snd{0.750} & \trd{0.327} 
\\
w/ $\cL_{bce}$, $G=6$ & \snd{21.53} & \fst{\textbf{0.753}} & \fst{\textbf{0.317}} 
\\
w/ $\cL_{bce}$, $G=1$ (Ours)  & \fst{\textbf{21.56}} & \fst{\textbf{0.753}} & \snd{0.321} \\
\bottomrule[1.2pt]
\end{tabular}
}
\vspace{-0.2cm} 
\caption{%
\textbf{Ablation on Voxel-to-3DGS VQ-VAE} on Waymo dataset. \textit{G} denotes to number of Gaussians per voxel.}
\label{tab:ablation_3d}

\end{table}
\begin{table}[t]
    \resizebox{0.44\textwidth}{!}{
 \begin{tabular}{l|cc|cc}
    \toprule[1.2pt]
     & \multicolumn{2}{c|}{KITTI-360}& \multicolumn{2}{c}{Waymo} \\
     & FID$\downarrow$  & KID$\downarrow$ 
    & FID$\downarrow$  & KID$\downarrow$  \\
    \midrule[1.2pt]
     Depth Cond. & 78.6 & 0.070 & 77.4 & 0.071 \\
     RGB Cond. (Ours)  & \fst{\textbf{36.9}}  & \fst{\textbf{0.026}}  & \fst{\textbf{41.3}}   & \fst{\textbf{0.030}}  \\ 
    \bottomrule[1.2pt]
  \end{tabular}
  }
  \vspace{-0.2cm} 
    \caption{
    \textbf{Ablation on Conditional Signals}. Both models are trained in the same way for the same number of steps.
  }
  \label{tab:ablation_2d}

\end{table}

\subsection{Ablation Study} 
\boldparagraph{Ablation on Voxel-to-3DGS VQ-VAE} As shown in~\cref{tab:ablation_3d}, we validate the effects of two components we use for training the Voxel-to-3DGS VQ-VAE. We compute PSNR, SSIM, LPIPS~\cite{lpips} on the Waymo~\cite{waymo} validation dataset. 
First, we ablate the effect of the BCE loss (w/o $\cL_{bce}$), which provides direct supervision to help our model learn scene geometry.
We find that while the VQ-VAE can still converge with only 2D supervision, it struggles to predict accurate scene occupancy, resulting in slightly degraded performance.
Next, we explore the effect of increasing the number of Gaussians per voxel from 1 to 6 ($G=6$) and find only modest improvements.
This matches observations in ~\cite{scaffoldgs,scube}.
A possible explanation is that predicting diverse Gaussians from a shared voxel feature is difficult in our generalizable model.
To balance GPU memory usage and performance, we set $G=1$ in our approach.

\begin{figure*}[!t]
    \centering
    \includegraphics[width=\linewidth]{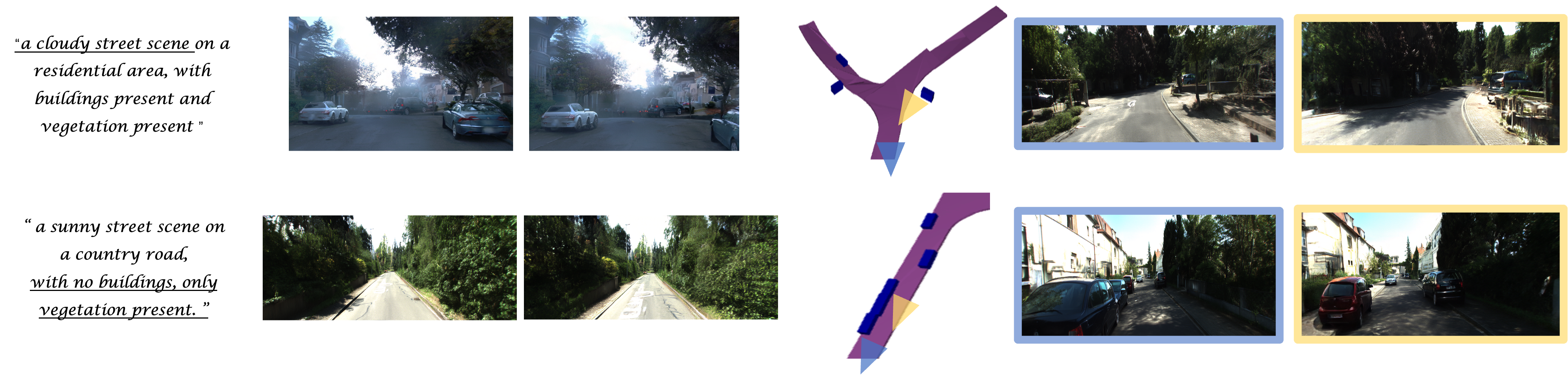} \\
     \vspace{-0.3cm}
    \caption{\textbf{Controllable Scene Synthesis on Waymo and KITTI360.} The visualization of conditional signals and corresponding synthesized images confirms the adherence of the generated content to diverse conditional guidance.}
    \label{fig:application}
\end{figure*}

\boldparagraph{Condition Signal for 2D Augmentation} 
For the ablation study of the 2D diffusion model, we utilize our SVD version for all experiments. We ablate our key design of generating coarse geometry and appearance in the 3D space.
Following another line of work \cite{infinicube, uniscene, lu2024wovogen, consistentcity, streetscapes} that purely consider appearance generation in the 2D space, we replace rendered RGB image as a rendered geometry buffer (depth map in our case) as condition and train our 2D video diffusion in the same manner. We evaluate final FID/KID~\cite{fid} in both cases and the results are shown in ~\cref{tab:ablation_2d}. Model conditions on the foreground RGB can synthesize images with higher quality at the same training steps since they provide direct texture information, hence improving training efficiency compared with using the indirect geometry conditions. We visualize different conditional signals and their synthesized full images in~\cref{fig:svd_ablation}. Under the guidance of the depth map, the surface of surrounding houses is much rougher and there are more noisy artifacts. Moreover, using depth as a conditioning signal struggles to preserve appearance consistency when revisiting the same place, whereas our RGB guidance enhances consistency by providing a coarse appearance guidance. More details about this experiment can be found in supplementary.

\begin{figure}[t]
    \centering
    \centering
   \setlength{\tabcolsep}{0.2pt}
   \small
  \begin{tabular}{ccc}
  \includegraphics[width=0.16\textwidth]{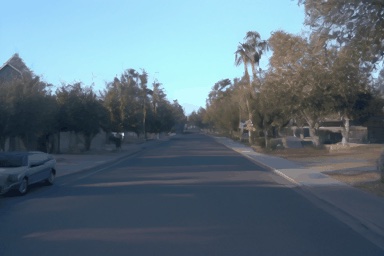} 
  &\includegraphics[width=0.16\textwidth]{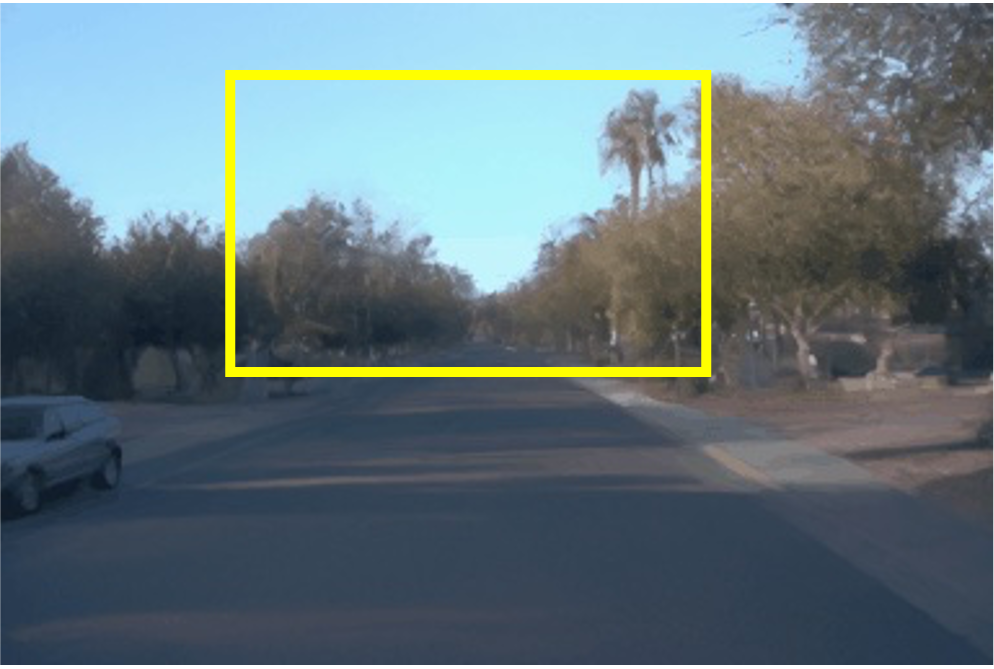} 
  &\includegraphics[width=0.16\textwidth]{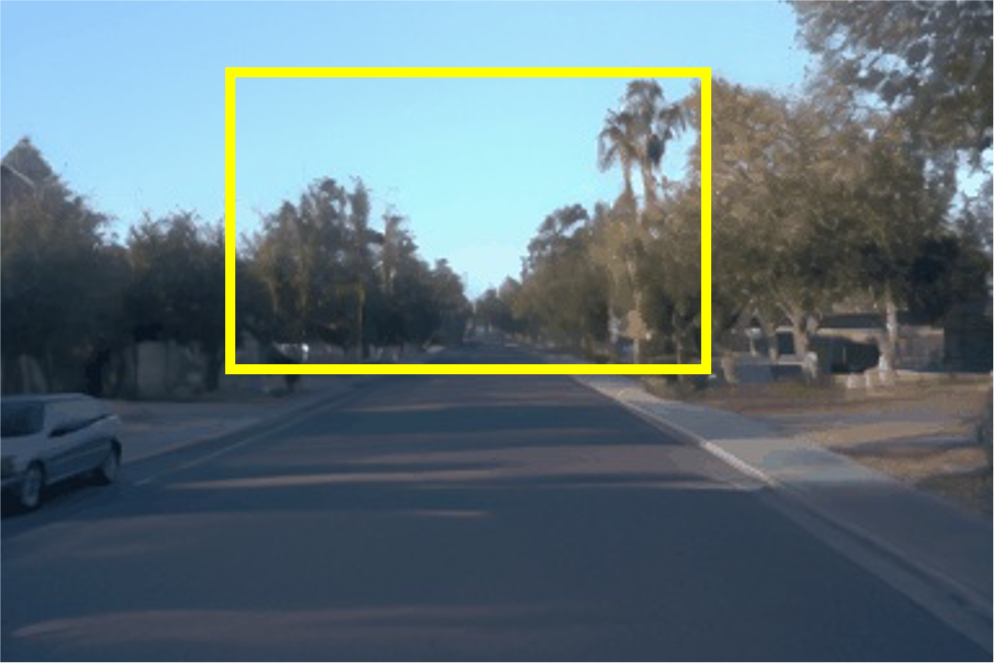} \\
   Pre. clip &  w/o DF &   w/ DF \\
  \end{tabular}
  \vspace{-0.2cm}
   \caption{\textbf{Ablation on Inference Strategy}. We visualize neighboring frames obtained from two clips. 
   Using the diffusion forcing strategy (w/ DF) significantly improves the consistency of background regions compared with the repaint strategy (w/o DF).
   \vspace{-0.2cm}
   }
    \label{fig:df_vis}

\end{figure}

\boldparagraph{Inference Strategy} As we only model the background with our video diffusion model and fine-tune it on clips with a maximum frame 5/17, we encounter sudden changes between clips for distant areas. To solve this problem,  we use previously generated frames as conditions for the later frames, inspired by ~\cite{diffusionforcing,chronodepth}. We show the impact of using the Diffusion Forcing strategy in ~\cref{fig:df_vis}. When using this strategy, frames between clips become consistent, otherwise the background will change drastically.

\begin{figure}[t]
    \centering
    \centering
\setlength{\tabcolsep}{1pt}
 \begin{tabular}{P{2cm}P{6cm}}
\includegraphics[width=0.3\columnwidth]{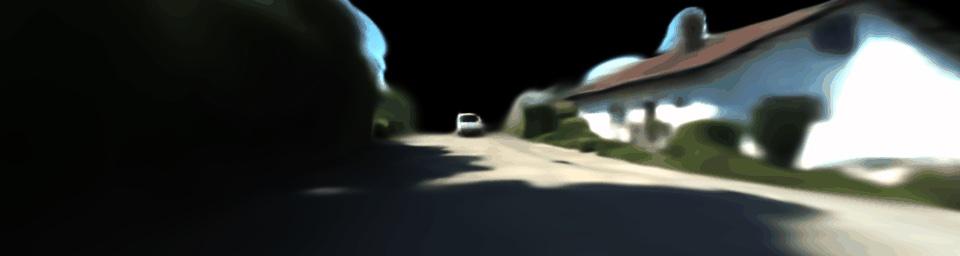} 
&\includegraphics[width=0.7\columnwidth]{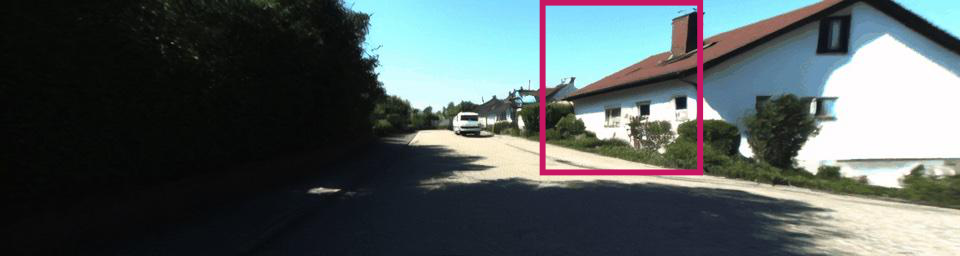} \\
\includegraphics[width=0.3\columnwidth]{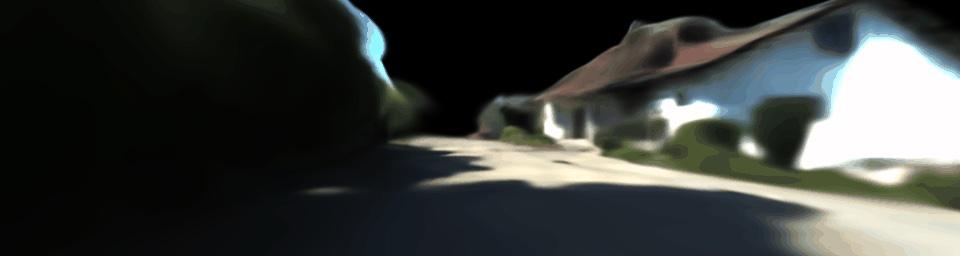} 
&\includegraphics[width=0.7\columnwidth]{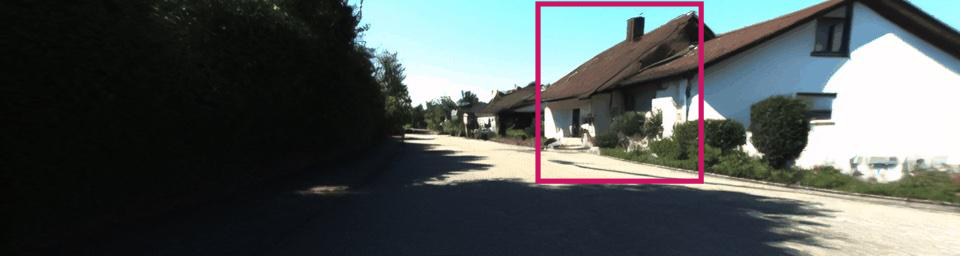} \\
\end{tabular}
 \vspace{-0.3cm}
\caption{\textbf{Inpaint Samples}. We use Repaint strategy~\cite{repaint} to inpaint original scenes, demonstrating our model's capability to generate diverse scenes.}
\label{fig:repaint}
\vspace{-0.5cm}

\end{figure}

\subsection{More Results}

\boldparagraph{Scene Controllability}To demonstrate the fine-grained controllability of our method during the generation process, ~\cref{fig:application} showcases scenes synthesized under different conditional signals. Conditioned on the textual instructions provide control over global attributes such as weather and background composition, for example specifying whether the scene should contain only vegetation or include buildings. In addition, given bounding boxes and road maps, our method can also generate scenes with diverse vehicle layouts and distinct road structures, such as straight roads and curved turns. 

\boldparagraph{Scene Inpainting} 
To showcase the generative capability of our model, we use the Repaint strategy~\cite{repaint} to re-generate certain parts of the original scene. Specifically, we fixed the first half of the scene latent, and inpaint another half part. It can be noticed in~\cref{fig:repaint} that the shape of the house in the latter part has been changed. %

\section{Conclusion} 
We propose \METHOD{}, a novel urban scene generator that leverages 3D-to-2D diffusion cascades. By introducing a novel 3D latent diffusion model and a refinement stage with a 2D diffusion model, our method improves image quality while maintaining strong camera and content controllability through the pure 3D generation stage, aided by diverse conditional signals. Our method highlights a promising direction for complex scene generation by integrating both diffusion paradigms to fully unlock their potential.\par
\boldparagraph{Limitations and Future Work}
One of our main challenges is that the quality of video diffusion depends on the preceding 3D LDM. Although the 2D refinement stage can mitigate some artifacts inherited from the 3D generation step, unsatisfactory 3D generation may cause corruption. Scaling up the training data and model size could further improve the visual quality of native 3D scene generation.

{
    \small
    \bibliographystyle{ieeenat_fullname}
    \bibliography{main}
}

\end{document}